%% file: egpaper_for_review.tex
\documentclass[10pt,twocolumn,letterpaper]{article}

\usepackage{iccv}
\usepackage{times}
\usepackage{epsfig}
\usepackage{graphicx}
\usepackage{amsmath}
\usepackage{amssymb}
\usepackage{algorithmic}
\usepackage{multirow}
\usepackage{makecell}
\usepackage[accsupp]{axessibility}
\usepackage[ruled,linesnumbered,noend]{algorithm2e} 

\usepackage[pagebackref=true,breaklinks=true,letterpaper=true,colorlinks,bookmarks=false]{hyperref}

\iccvfinalcopy 

\def\iccvPaperID{6859} 

\ificcvfinal\fi

\begin{document}

\title{Auto Graph Encoder-Decoder for Neural Network Pruning}
\author{Sixing Yu \\
Iowa State University\\
{\tt\small yusx@iastate.edu}
\and
Arya Mazaheri\\
Technical University of Darmstadt\\
{\tt\small arya.mazaheri@tu-darmstadt.de}

\and
Ali Jannesari\\
Iowa State University\\
{\tt\small jannesari@iastate.edu}
}

\maketitle
\ificcvfinal\thispagestyle{empty}\fi
\input{00_abstract}
\input{01_Introduction}
\input{02_RelatedWork}

\input{03_Methodology}

\input{04_Experiment}
\input{05_Conclusion}

{\small
\bibliographystyle{ieee_fullname}
\bibliography{egbib}
}
\end{document}

%% file: 00_abstract.tex
\begin{abstract}

Model compression aims to deploy deep neural networks (DNN) on mobile devices with limited computing and storage resources. However, most of the existing model compression methods rely on manually defined rules, which require domain expertise. DNNs are essentially computational graphs, which contain rich structural information. In this paper, we aim to find a suitable compression policy from DNNs' structural information. We propose an automatic graph encoder-decoder model compression (AGMC) method combined with graph neural networks (GNN) and reinforcement learning (RL). We model the target DNN as a graph and use GNN to learn the DNN's embeddings automatically. We compared our method with rule-based DNN embedding model compression methods to show the effectiveness of our method. Results show that our learning-based DNN embedding achieves better performance and a higher compression ratio with fewer search steps.
We evaluated our method on over-parameterized and mobile-friendly DNNs and compared our method with handcrafted and learning-based model compression approaches. On over parameterized DNNs, such as ResNet-56, our method outperformed handcrafted and learning-based methods with $4.36\%$ and $2.56\%$ higher accuracy, respectively. Furthermore, on MobileNet-v2, we achieved a higher compression ratio than state-of-the-art methods with just $0.93\%$ accuracy loss.
\end{abstract}

%% file: 01_Introduction.tex
\section{Introduction}

With the increasing demand to deploy deep neural networks~(DNNs) on edge devices (e.g., mobile phones, robots, self-driving cars, etc.), which usually have limited storage and computing power, model compression techniques became essential for efficient DNN deployment. Network pruning~\cite{han2015deep,hassibi1992second,molchanov2019importance}, factorization~\cite{swaminathan2020sparse,2013low-rank}, knowledge distillation~\cite{polino2018model,park2019relational,hinton2015distilling}, and parameter quantization~\cite{han2015deep,wang2020bamc,jacob2018quantization} are among the most well-known model compression techniques. 
However, these methods heavily rely on handcrafted rules defined by experts, demanding an extensive amount of time and might not necessarily lead to a fully compressed model.

Recently, automatic model compression~\cite{he2018amc,wang2020bamc,liu2020autocompress} has gained momentum. For example, Wang et al.~\cite{wang2020bamc} proposed a Bayesian automatic model compression method trained in a one-shot manner to find reasonable quantization policies. He et al.~\cite{he2018amc} proposed an automatic model compression method based on reinforcement learning (RL). However, when representing DNNs, they rely on manually defined DNN embedding vector~(e.g., using one-hot vectors to characterize DNN's hidden layers) and ignore the rich structural information between the hidden layers.

DNNs are essentially represented as computational graphs in deep learning frameworks, such as TensorFlow~\cite{abadi2016tensorflow} and PyTorch~\cite{paszke2019pytorch}. A computational graph is composed of numerous primitive operations (e.g., add, minus, multiply), where edges are operations and nodes are intermediate calculation results (i.e., feature maps in DNNs).
Such a rich structural representation can effectively delineate the state of DNN hidden layers. 
Additionally, computational graphs often contain repetitive structural patterns due to the same set of primitive operations being used multiple times. Thus, we aim to benefit from this feature by extracting the structural information readily available within computational graphs to identify the redundancy and pruning policy for DNN hidden layers. 

In this paper, we propose a graph-based Auto Graph encoder-decoder Model Compression (AGMC) method that combines graph convolutional networks (GCNs)~\cite{article,NEURIPS2018_53f0d7c5,xu2019how} and reinforcement learning (RL) ~\cite{lillicrap2015continuous,sutton1999policy,DPG} to learn the compression strategy of DNNs without expert knowledge. 
The graph encoder-decoder aims to learn the DNN's layer embeddings. The GCN-based graph encoder learns the DNN representation from its structure information, and the decoder decodes the representation to hidden layer embeddings. The RL agent takes the hidden layer embeddings as the environment states, looks for the pruning ratio for each hidden layer, and generates a corresponding compressed candidate model. Finally, we evaluate the candidate compressed model performance and provide a reward value as feedback to the RL agent.
By exploiting DNN's structure information to suggest compression policies, our approach successfully applied network pruning and achieved outstanding results on various DNNs, such as ResNet~\cite{he2016deep}, VGG-16~\cite{Simonyan2015VGG}, MobileNet~\cite{Andrew2017MobileNetv1,Sandler2018mobileNetv2}, and ShuffleNet~\cite{zhang2018shufflenet,ma2018shufflenetv2}.

In essence, this paper makes the following contributions:
\begin{itemize}
    \item A novel automatic layer embedding based on computational graph's structure.
    \item An efficient method based on GCN and RL to automate the channel pruning.
    \item State-of-the-art model pruning results on various DNN models.
\end{itemize}






%% file: 02_RelatedWork.tex
\section{Background and related work}
With the increasing demand to make edge devices smarter using AI, efficient deep neural network design is more important than ever. Hence, various efficient networks have been introduced to reduce the computational complexity and memory requirement of such networks. MobileNet-v1/v2~\cite{Andrew2017MobileNetv1,Sandler2018mobileNetv2}, ShuffleNet-v1/v2~\cite{zhang2018shufflenet,ma2018shufflenetv2}, DiCENet~\cite{mehta2020dicenet}, and CondenseNet~\cite{huang2018condensenet} are among the notable efforts that introduced custom convolutional blocks to improve the overall efficiency.
Furthermore, neural architecture search (NAS)~\cite{zoph2017neural,gao2019graphnas,chen2019renas,mammadli2019effdnn} methods also attempt to generate efficient DNNs by searching for the most optimal neural network structure, given the constraints of the target hardware platform.

Within the context of this paper, we discuss former studies related to model compression, particularly network pruning and the application of GCN in model compression. In the following, we will provide a brief overview of these methods.

\textbf{Model compression.} A multitude of previous work focus on model compression techniques, such as knowledge distillation~\cite{polino2018model,park2019relational,hinton2015distilling}, parameter quantization~\cite{han2015deep,wang2020bamc,jacob2018quantization}, factorization~\cite{swaminathan2020sparse,2013low-rank}, and network pruning~\cite{han2015deep,hassibi1992second,molchanov2019importance}.

As DNNs are typically over-parameterized, network pruning is among the most widely used model compression techniques, which has achieved outstanding results and can dramatically shrink model size~\cite{blakeney2020is}.
It eliminates a portion of parameters and computation within each DNN layer via two different methods: (1) fine-grained pruning and (2) structured pruning. Fine-grained pruning~\cite{han2015deep} targets individual unimportant elements in weight tensors. On the other hand, structured pruning~\cite{FP_li2017pruning} attempts to prune entire blocks of weight tensors, such as channels, rows, columns, and blocks.
Although the fine-grained pruning could achieve a high compression rate with minimal accuracy loss, they lead to irregular sparsity patterns, demanding specialized hardware accelerators~\cite{ji2018recom,han2017ese} to realize any speedup.
Alternatively, using structured pruning results in regularly pruned weights and can be used on commodity hardware. In this paper, we particularly focus on structured pruning.

Empirical pruning policies are uniform, shallow, and deep~\cite{he2017channel,FP_li2017pruning}.
The uniform policy sets the compression ratio uniformly. The shallow and deep policies aggressively prune shallow and deep layers, respectively. Such handcrafted empirical policies heavily rely on manually defined rules and might not lead to an optimal compression policy. 
Other handcrafted methods focusing on channel pruning are SPP\cite{wang2017SPP}, FP~\cite{FP_li2017pruning}, and RNP~\cite{Lin2017RNP}. 
SPP prunes DNNs by analyzing each layer and measures the reconstruction error to determine the pruning ratios. FP evaluates the performance of single-layer pruning and estimates the sensitivity of each layer. Layers with lower sensitivity are pruned more aggressively. RNP introduced an RL-based method and groups all convolutional channels into four sets for training. 

Conventional network pruning methods primarily rely on handcrafted and rule-based policies, demanding human effort and domain expertise.
Moreover, such methods might not necessarily offer a fully compressed model. 
Recently, RL-based automatic network pruning methods~\cite{xiao2019autoprun,he2018amc,liu2020autocompress} have been proposed. Liu et al.~\cite{liu2020autocompress} proposed an ADMM-based~\cite{ADMM} structured weight pruning method and an innovative additional purification step for further weight reduction. He et al.~\cite{he2018amc} proposed AutoML for network pruning, which leverages RL to predict compression policies, yet they still use handcrafted rules (11 fixed features) to represent DNNs and ignore the rich structural information within computational graphs. 

\textbf{Graph neural networks.} 
GNN and its variants~\cite{kipf2017gcn,Schlichtkrull2018rgcn} are successfully applied to learn the topology information from graphs. For instance, they have been successfully applied to node classification, link prediction, and graph classification.
Furthermore, graph-based NAS methods~\cite{Guo2019NAS_NAT,Han2020NAS_oneshot,Dudziak2021BPR_NAS} model DNNs as computational graphs and find the optimal DNN structure from a graph-based search space. These methods inspired us to use a GCN-based graph encoder to learn the DNN embeddings.


%% file: 03_Methodology.tex
\input{Figure_agmc_overview}
\section{Methodology}

To prune a given DNN, we first modeled the DNN as a computational graph and introduced a GCN-based graph encoder to learn the DNN's representation $g$.
Then the decoder decodes the $g$ in to layer embeddings $s_i \in S, i=1,2,..,T$, where $T$ is the number of hidden layers. 
Since we aim to compress the DNN by predicting the pruning ratio for each hidden layer, the RL agent takes the layer state $S$ as the environment state to search for the hidden layer's pruning policy $a_i \in A, i=1,2,..,T$. The pruned DNN's performance is then used as a reward for the current actions $A$ taken by the RL agent.\\
Figure \ref{fig:agmc_overview} depicts an overview of our method. 
In the following, we will explain the details of the simplified computational graph, graph encoder-decoder, and RL agent within our approach.

\subsection{Simplified graph representation of DNNs}
The computational graph representation of DNNs is composed of numerous primitive operations (e.g., add, minus, multiply), where edges are operations and nodes are intermediate results (i.e., feature maps).
Thus, a typical computational graph might involve billions of primitive operations~\cite{he2016deep}, making it unrealistic to use the graph for our analysis directly.
To simplify the graph representation, we choose commonly used machine learning operations as primitive operations $\mathcal{O} = $\{$n~\times~n$ conv, Relu, BatchNorm, (Max/Average) Pooling, Padding, Splitting\}. 
Such a simplification can significantly reduce the graph complexity and yet preserve important structural information.

\input{Figure_computational_graph}
Formally, We model a given DNN as a single-source and single-sink computational graph $G = (V,E,\mathcal{O})$, where $V$ is the node-set, $E$ is the edge set, and $\mathcal{O}$ is the primitive operation set. Each directed edge with an edge type is associated with a primitive operation in $\mathcal{O}$.
Figure~\ref{fig:computational_graph}~(a) shows the idea behind the simplified computational graph using two primitive operations $\mathcal{O} = $ \{1$\times$1 conv, 3$\times$3 conv\}, which correspond to two edge types. The computation graph $G$ denotes a compound operation composed of the primitive operations in $\mathcal{O}$:
\begin{equation}
y = assemble(conv3(conv1(x)),conv3(conv1(x)))
\end{equation}

Figure~\ref{fig:computational_graph}~(b) shows another example for constructing a computational graph for a ResNet block, which contains a 1$\times$1 convolutional layer with four output channels and a 3$\times$3 convolutional layer with three output channels. Although different layers have different computational graphs, they often share similar structures. 

\subsection{Auto graph encoder-decoder}

We introduce a GCN-based graph encoder-decoder to learn the embeddings of the target DNN's hidden layers automatically. The GCN-based graph encoder embeds the graph and learns the DNN's structure representation $g\in\mathbb{R}^{1 \times d}$, where $d$ is the embedding size. We also introduced an LSTM~\cite{LSTM} based decoder that decodes the DNN's representation to layer embeddings $S\in\mathbb{R}^{T \times d}$, where $N$ is the number of hidden layers.

\subsubsection{GCN-based graph encoder}
The GCN embeds graphs by aggregating node features from neighbor nodes. The message passing function can be formulated as follows:
\begin{equation}
    h^{l+1}_i = \sum_{j\in N_i}\frac{1}{c_i}W^l h^l_j,
\end{equation}
where $h^{l}_i$ is the hidden state of $i^{th}$ node in GCN's $l^{th}$ convolution, $c_i$ is a constant coefficient, $N_i$ is node $i$ neighbors, and $W^l$ is GNN's learnable weight matrix.

Although standard GCN and its variants aim to learn the node embeddings in a graph, we aim to learn the entire graph representation. Thus we need to take the graph representation from the node embeddings. 
One of the most commonly used mechanisms to achieve this is to use the graph mean pool~(Equation~\ref{eq:meanpool}), which averages the node embeddings.
The graph encoder is formulated as Equation~\ref{eq:H_GCN}. It embeds the computational graph and gets the node-embedding matrix $H$. Then, the graph mean pool reads the graph representation $g$ from the node embeddings.

\begin{equation}
\begin{aligned}
H = \text{GCN}_{encoder}(G) \in \mathbb{R}^{N \times d},
\label{eq:H_GCN}
\end{aligned}
\end{equation}

\begin{equation}
\begin{aligned}
g = \frac{1}{N} \sum_{i=1}^{N} {h}_i,
\label{eq:meanpool}
\end{aligned}
\end{equation}
where $H={h}_i, i=1,2,...,N$ is the node-embedding matrix, ${h}_i$ is the embedding of $i^{th}$ node , $N$ is the total number of nodes in the graph, and $d$ is the embedding size.

\subsubsection{Decoder}
The decoder aims to learn the environment states of DNN hidden layers for the RL agent.
Since the state vectors in the RL environment are determined by the previous state and the action (the pruning ratio), the decoder takes the previous layer’s states vector and RL agent's action as input:
\begin{equation}
    s_1=\text{LSTM}_{decoder}\left(g\right),
\end{equation}
\begin{equation}
    s_t=\text{LSTM}_{decoder}\left(s_{t-1},a_{t-1}\right)
\end{equation}
For the $t-$th hidden layer, we use the feature $s_{t-1}$ of the previous hidden layer and the compression policy $a_{t-1}$ (the action selected by the RL agent) to calculate the environment states.

\subsection{Automatic network pruning using reinforcement learning}
We leverage reinforcement learning to find the optimal pruning ratios efficiently. In the following, we describe the details of our reinforcement learning setup.
\paragraph{Environment states.}
In contrast to existing RL-based model compression methods that use fixed handcrafted layer embeddings as environment states, we use DNN layer embeddings $S\in\mathbb{R}^{T\times 1 \times d}$ generated by the graph encoder-decoder as environment states.

\paragraph{Action space.}
The actions made by the RL agent are pruning ratios within a continuous space. Specifically, the RL agent takes the layer embeddings $S\in\mathbb{R}^{T \times d}$ as environment states and predicts corresponding pruning ratios $a_i \in A, i=1,2,..,T$, where $a_i\in\left[0,1\right)$.

\paragraph{Reward function.}
We prune the DNN according to the pruning ratio made by RL agent, and use the performance of the compressed model as the reward. The reward function is defined in Equation~\ref{eq:reward}.
\begin{equation}
    R_{err} = -Error,
    \label{eq:reward}
\end{equation}
where \textit{Error} is the compressed DNN's top-1 error on the validation set.

\paragraph{Deep deterministic policy gradient (DDPG).}
Various RL policies aim to search within a continuous action space, such as proximal policy optimization~(PPO)~\cite{schulman2017ppo} and deep deterministic policy gradient~(DDPG)~\cite{lillicrap2015continuous}.
Similar to the AMC~\cite{he2018amc} method, we opted for DDPG as the RL policy to make a fair comparison and exclude the influence of RL policy on the experimental results. This way we can demonstrate the superiority of our learning-based embedding compared to the handcrafted rules.

The DDPG agent's search process can be formulated as following:
\begin{equation}
    g \in \mathbb{R}^{1 \times d} = GraphEncoder({G}),
    \label{eq:graphencoder}
\end{equation}
\begin{equation}
    S \in \mathbb{R}^{T\times 1\times d} = Decoder({g}),
    \label{eq:decoder}
\end{equation}
\begin{equation}
    A \in \mathbb{R}^{T\times 1\times 1} = MLP(S), 
    \label{eq:mlp}
\end{equation}
where $G$ is the computational graph, $g$ is the graph representation, $S$ is the environment states, and MLP is a multi-layer perceptron neural network.
The graph encoder embeds the graph $G$ and learns the DNN representation $g$ and the decoder decodes $g$ into hidden layers embeddings $s_i \in S, i=1,2,..,T$. Finally, the RL agent takes $S$ as environment states and uses MLP to project the embedding into hidden layer pruning ratios $a_i \in A, i=1,2,..,T$.

\subsubsection{Action rescaling}
The reward function that we use offers small or no incentive for model size and FLOPs reduction. Without any constraint~(e.g., FLOPs or \#parameters), the RL agent tends to search for a tiny compression ratio. Thus, to obtain the desired model size reduction, we apply Algorithm~\ref{alg} to adjust the action space $a$. 

In essence, Algorithm~\ref{alg} computes the size we still needed to reduce according to the original scale. Lines 1-2 compute the total model size~(e.g., FLOPs and \#parameters) $W_{all}$ and reduced size $W_{reduced}$. If the reduced size is less than the desired model size reduction $d$, the algorithm will rescale the pruning ratios to compensate the difference $d - W_{reduced}$. Lines 4-7 relate to the rescaling process, and the for-loop in lines 5-7 adjusts the pruning ratio for each layer according to the difference to the desired model size reduction. 
Finally, in line 7, we truncate the pruning ratio with the upper bound $a_{max}$.

\input{Algorithm_1}

%% file: Figure_agmc_overview.tex
\begin{figure*}
\begin{center}
   \includegraphics[width=1\linewidth]{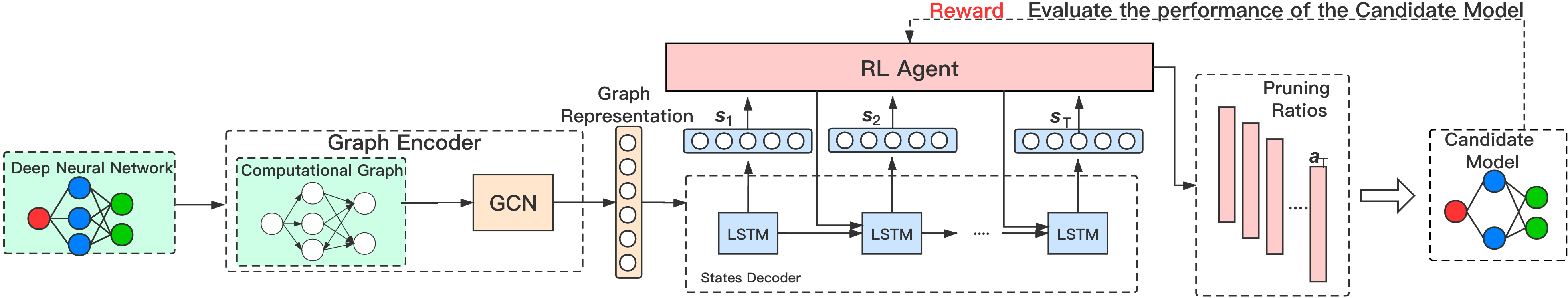}

\end{center}
    \caption{The workflow of  Auto  Graph encoder-decoder Model Compression (AGMC) }
\label{fig:agmc_overview}
\end{figure*}

%% file: Figure_computational_graph.tex
\begin{figure*}
\begin{center}
   \includegraphics[width=0.9\linewidth]{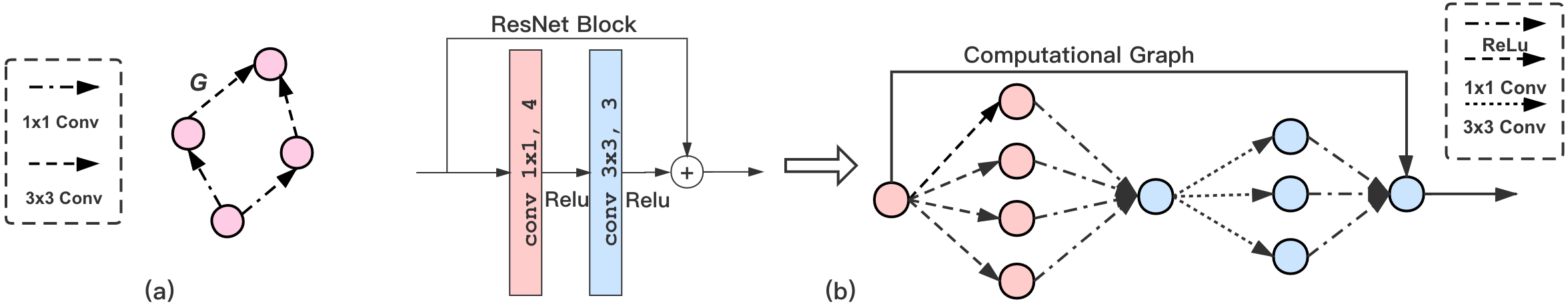}

\end{center}
   \caption{Simplified computational graph. (a)~An example of a simplified computational graph. (b)~Construction of a simplified computational graph for a ResNet block~\cite{he2016deep}}
\label{fig:computational_graph}
\end{figure*}

%% file: Algorithm_1.tex
\begin{algorithm}

\KwIn{The actions $a=\{a_0,...,a_T\}$, the upper bound of actions $a_{max}$, the model size (\#FLOPs/\#Parameters etc.) of each hidden layer $W = \{W_0,...,W_T\}$, and the desired model size reduction $d$}
\KwOut{The actions $a'$ after re-scaling}
$W_{all} = \sum_t W_t$\\
$W_{reduced} = \sum_t W_ta_t$\\
\If{$W_{reduced} < d$} {
		$d_{rest} = d - W_{reduced} $\\
		\For{$i = 1,2,...,T$}{
			$a_i += (d_{rest} *( a_i / \sum_t a_t))/ W_i$\\
			$a'_i = min(a_{max},a_i) $
		}
	}

 {\Return $a'$}
\caption{Rescaling the actions for the desired model size reduction}
\label{alg}
\end{algorithm}

%% file: 04_Experiment.tex
\section{Experimental results}
We evaluate our approach by performing FLOPs-constrained structured pruning on several convolutional networks, including over-parameterized DNNs (e.g., ResNet-20/56~\cite{he2016deep} and VGG-16~\cite{Simonyan2015VGG}) and mobile-friendly DNNs~(e.g., MobileNet-v1/v2~\cite{Andrew2017MobileNetv1,Sandler2018mobileNetv2} and ShuffelNet-v1/v2~\cite{zhang2018shufflenet,ma2018shufflenetv2}).
To show the superiority of our approach, we compared our approach with various existing methods in different categories, such as:
\begin{itemize}
    \item Uniform, shallow, and deep empirical policies~\cite{he2017channel,FP_li2017pruning}.
    \item Handcrafted channel reduction methods, such as SPP\cite{wang2017SPP}, FP~\cite{FP_li2017pruning}, and RNP~\cite{Lin2017RNP}.
    \item Regularization-based methods, such as MorphNet~\cite{gordon2018morphnet} and SSL~\cite{wen2016ssl}.
    \item RL-Based AutoML methods, such as auto model compression method AMC~\cite{he2018amc}, which manually defines DNN layer embeddings, and random search with reinforcement learning~(RS), which does not leverage any layer embeddings.
    \item Other pruning methods, such as DSA~\cite{vedaldi2020dsa} and Rethink~\cite{liu2019rethinking}. 
\end{itemize}

Finally, we show the inference acceleration and memory saving of compressed models on a GPU platform.

\paragraph{RL setup.} The actor network $\mu$ and the critic network $Q$ have two hidden layers, each with 300 units. The $\mu$'s output layer applies the sigmoid function to bound the actions within $(0,1)$. We use $\tau = 0.01$ for the soft target updates. In the first 25 episodes, our agent searches with random action. Then, it continues searching for 300 episodes with exponential decayed noise. The graph encoder is a two-layer GCN with a hidden feature size of 50 units and a DNN embedding size of 11 units.

\paragraph{Datasets.} We conducted our experiments using CIFAR-10~\cite{krizhevsky2009cifar}, CIFAR-100~\cite{krizhevsky2009cifar} and ILSVRC-2012~(ImageNet)~\cite{olga2015ImageNet} datasets. To accelerate the search process on CIFAR-10/100, we split the training set to $15K$ and $5K$ images. We used the $15K$ training set to rapidly fine-tune the candidate model and the remaining $5K$ images as the validation set to calculate the reward function. In the ILSVRC-2012 dataset, we split $5K$ images from the training set as the validation set to calculate the reward. 
The Validation accuracy of the ILSVRC-2012 dataset is very sensitive to the compression, as with high compression ratios, the accuracy drops considerably without fine-tuning. Thus, the RL agent can not get a valuable reward.
As a remedy, we decompose the pruning on the ILSVRC-2012 dataset into several stages and add one epoch of fine-tuning for each search episode. For instance, to obtain a 49\% FLOPs model compared to the original network, instead of performing a single step 49 \% FLOPs pruning, we prune the target DNN two times, each with 70\% FLOPs constraint~(i.e., 70\%FLOPs~$\times$ 70\%FLOPs = 49~\% FLOPs).

\subsection{The effectiveness of DNN embeddings}
In contrast to existing methods~\cite{liu2020autocompress,he2018amc}, layer embeddings are essential for our learning-based automatic network pruning method. In the following, we analytically compare the effectiveness of using our DNN graph embedding with existing approaches.

\paragraph{Learning-based vs. manually-defined layer embeddings.} We compare AGMC with AMC~\cite{he2018amc}, which manually defines 11 features related to each layer as the embedding vector $s_t=(t,n,c,h,w,stride,k,FLOPs(t),reduced,rest,a_{t-1})$
, where $t$ is the layer id, the dimension of the kernel is $n\times c\times k\times k$, and the input is $c\times h\times w$. $FLOPs(t)$ is the FLOPs of layer $t$. $Reduced$ is the total number of FLOPs reduced in previous layers and finally $rest$ is the number of remaining FLOPs in the layers ahead. 
We argue that such a strict layer embedding may miss important information, such as the number of parameters in each hidden layer, which are only applicable to a given DNN. 
In the AGMC, the graph encoder-decoder learns the layer states from DNN structural information. Thus, it does not require expert knowledge and is applicable for all kinds of DNNs.

Since AMC has defined 11 features to represent a convolutional layer, we also set the learning-based embedding size to $11$ (i.e., $S\in\mathbb{R}^{T \times 11}$). We evaluated our learning-based embeddings and the AMC handcrafted embedding on ResNet-20 pre-trained with the CIFAR-10 dataset.
Figure \ref{fig:layer_embedding} shows the spatial decomposition evaluation for the AMC's layer embeddings under 50\% FLOPs ResNet-20. 
Using stride as the only layer embedding, we get an error rate of $31\%$ since it is difficult to distinguish different layers. However, combining the stride with the number of filters $n$ decreases the error rate to $18\%$. Consequently, combining all the features leads to $10.2\%$ error margin. On the other hand, our learning-based layer embedding achieves an error rate of $5.38\%$, outperforming the manually defined layer embedding by a factor of two.
\input{Figure_4}

\paragraph{Learning-based vs. no embedding.} 
We compared AGMC with random search~(RS) without layer embedding. We set all the hidden layers of the RS setup to a fixed one-hot vector as the RL agent's environment state and use $R_{err}$ as a reward. 
Then, we leverage the DDPG reinforcement learning agent to search pruning ratios for ResNet-20/56. ResNets have residual connections between their blocks, which instructs equal channel size between residual connection blocks. We opted for removing all the residual connections to avoid sharing the pruning ratios between residual connected layers and learn each hidden layer's embedding independently. 
As shown in Figure \ref{fig:random_search}, AGMC achieves better results compared with RS on ResNet-20. 
Particularly, AGMC enabled us to find the compressed model with fewer episodes, higher accuracy, and more FLOPs reduction. Moreover, by further layer-wise analysis, we observed that AGMC tends to prune each layer uniformly and the pruning ratio is more stable than the RS. Such an observation is in line with the uniform pruning policy\cite{FP_li2017pruning}, which argues that the uniform policy can yield better pruning. 

Additionally, we pruned ResNet-56 under different FLOPs-constraints. ResNet-56 contains 56 convolutional layers, which is far deeper and more challenging than ResNet-20 to prune. Figure~\ref{fig:Acc_FLOPs} depicts the validation accuracy under different pruning ratios.
In all cases, AGMC outperforms RS, as more FLOPs are pruned. For instance, with 10\% FLOPs reduction, the performance of AGMC and RS are almost the same. However, with 90\% FLOPs reduction, the AGMC surpasses the RS by a large margin.

\input{Figure_5}
\input{Figure_Acc_FLOPs}

\paragraph{Generalizability of the graph encoder.}
AGMC adopts a GCN-based graph encoder to embed DNNs topology structure. Since we model the DNNs as graphs under the same rule, the graph encoder trained on one DNN should also achieve similarly good performance on other similar DNNs. 
Thus, we investigated whether AGMC has learned the structural pattern of ResNet-56 while performing channel pruning. We transferred the trained AGMC to ResNet-20, which is a similar network. 
When searching the pruning ratio of ResNet-20, we only updated the decoder parameters and did not require the graph encoder and RL agent's gradients. 
With 100 search episodes and $50\%$ FLOPs reduction on ResNet-20, the result of AGMC transferred from ResNet-56 and obtained the validation accuracy of $92.08\%$, which is similar to the AGMC trained on ResNet-20 with $94.6\%$ validation accuracy.


\subsection{Over-parameterized DNN pruning}
We evaluate AGMC on ResNet-20/32/44/56/110 and VGG-16, often considered as over-parameterized DNNs. Such deep and compact networks involve billions of parameters, incurring high memory consumption. Thus, it is challenging to deploy them on edge devices with limited computing and power budgets.

We perform FLOPs-constrained pruning on over-parameterized DNNs by leveraging the RL agent to search for pruning ratios for each convolutional layer. However, ResNet has residual connections, and different pruning ratios between residual connected layers will lead to feature-map dimension mismatch. To overcome this issue, we share the pruning ratio between the residual connection layers. Additionally, we follow the same experiment settings as in DSA~\cite{vedaldi2020dsa} since it has a significant impact on the pruning results. For instance, the number of fine-tuning epochs is one of the key factors, where a larger value leads to higher test accuracy but with the cost of additional time and resources. 

Table~\ref{table:1} reports the results of AGMC in comparison with existing pruning methods for over-parameterized networks.
Our method outperforms the empirical policies~\cite{he2017channel,FP_li2017pruning} by a large margin with $7.42\%$ higher accuracy on ResNet-20 and $4.36\%$ on ResNet-56.
Compared with the RL-based method, AMC~\cite{he2018amc}, AGMC achieved $5.02\%$ and $2.56\%$ higher accuracy on ResNet-20 and ResNet-56, respectively. 
Moreover, the networks pruned by AGMC yielded less accuracy loss compared with rule-based pruning methods~\cite{gordon2018morphnet,liu2019rethinking,he2018sfp,vedaldi2020dsa}.
Additionally, we recorded the RL search time on ResNet-56 with 300 episodes. It takes $(320\pm30)$ seconds on RTX 8000 GPU to finish the entire search for the pruning ratios.
For VGG-16 model trained on the ILSVRC-2012 dataset, we compared AGMC with handcrafted channel reduction methods (i.e., SPP~\cite{wang2017SPP}, FP~\cite{FP_li2017pruning}, and RNP~\cite{Lin2017RNP}) and AMC~\cite{he2018amc}.
Results show that AGMC outperformed all the baselines methods by a large margin. 

\input{Table_1}

\subsection{Mobile-friendly DNN pruning}
We further evaluated AGMC on mobile-friendly DNNs, such as MobileNet-v1/v2~\cite{Andrew2017MobileNetv1,Sandler2018mobileNetv2} and ShuffleNet-v1/v2~\cite{zhang2018shufflenet,ma2018shufflenetv2}. 
Instead of using standard convolutional layers, mobile-friendly DNNs have designed customized convolutional blocks to reduce the parameters, leading to better performance on edge devices. 
For instance, the MobileNet-v1 block splits a traditional convolution into a pair of point-wise and depth-wise convolutions. Based on MobileNet-v1, MobileNet-v2 adds an additional linear expansion layer and introduced residual connections.
To maintain the characteristics of the mobile-friendly DNNs, we have developed specific pruning strategies for them. 

\paragraph{MobileNet-v1.} The MobileNet-v1 block contains a depth-wise and a point-wise convolution, instead of pruning them separately, we view the two convolutions together and only prune point-wise convolutions. Since the depth-wise convolution only operates on one input channel and pruning that filter will cause information loss for the corresponding channel.

\paragraph{MobileNet-v2.} Similar to MobileNet-v1, we prune linear expansion layers and point-wise convolutional layers. Since residual connections are between linear expansion layers, we share the linear expansion layers' pruning ratio.

\paragraph{ShuffleNet-v1/v2.}
ShuffleNet uses blocks containing depth-wise and point-wise convolutions, channel shuffle, linear expansion, and residual connections. To avoid dimension mismatch when downsampling, we consider the ShuffleNet blocks together and perform channel pruning inside the blocks. In a ShuffleNet block, we do not prune the expansion layer (the output layer of the block) and only prune the point-wise filters.

\input{Table_2}

The results for mobile-friendly networks are given in Table \ref{table:2}.
On MobileNet-v1/v2, we compare AGMC with the uniform pruning policy and the RL-based method AMC. 
Compared to the uniform policy, which sets the compression ratio uniformly, AGMC achieves a higher compression ratio with only $1.2\%$ test accuracy loss. 
Furthermore, our efficient layer embedding outperforms AMC on both MobileNet-v1 and MobileNet-v2, with the same target FLOPs. 
Similarly, our method succeeded in pruning 40\% of the ShuffleNet-v1/v2 FLOPs and obtaining a more accurate compressed model than random search.

\input{Table_3}

\subsection{Inference acceleration and memory saving}
Here, we discuss the inference speed of the compressed ResNet-20/56, VGG-16, and MobileNet-v1 on an Nvidia RTX 2080Ti GPU. AGMC performs channel pruning on convolutional layers, accelerating the inference on parallel devices like GPUs. We calculated the inference speed of the pruned models and compared them with the original model. We used batch size 32, and the compressed models are tested on CIFAR-10 and ILSVRC-2012 datasets.
As shown in Table \ref{table:3}, the models pruned by AGMC achieve notable GPU memory reduction. For example, for VGG-16, the original model's GPU memory usage is 528~MB, since it has dense layers and its first dense layer contains 25088 neurons. 
The $20\%$ FLOPs VGG-16 with pruned convolutional layers significantly reduced the feature map size input to dense layers, taking 141~MB memory less than the original. Moreover, without losing too much test accuracy, all the models pruned by AGMC achieved remarkable inference speedup. For instance, the $20\%$ FLOPs VGG-16 achieved $1.22 \times$ speedup on the ImageNet dataset.



%

%% file: Figure_4.tex
\begin{figure}[t]
\begin{center}
   \includegraphics[width=1\linewidth]{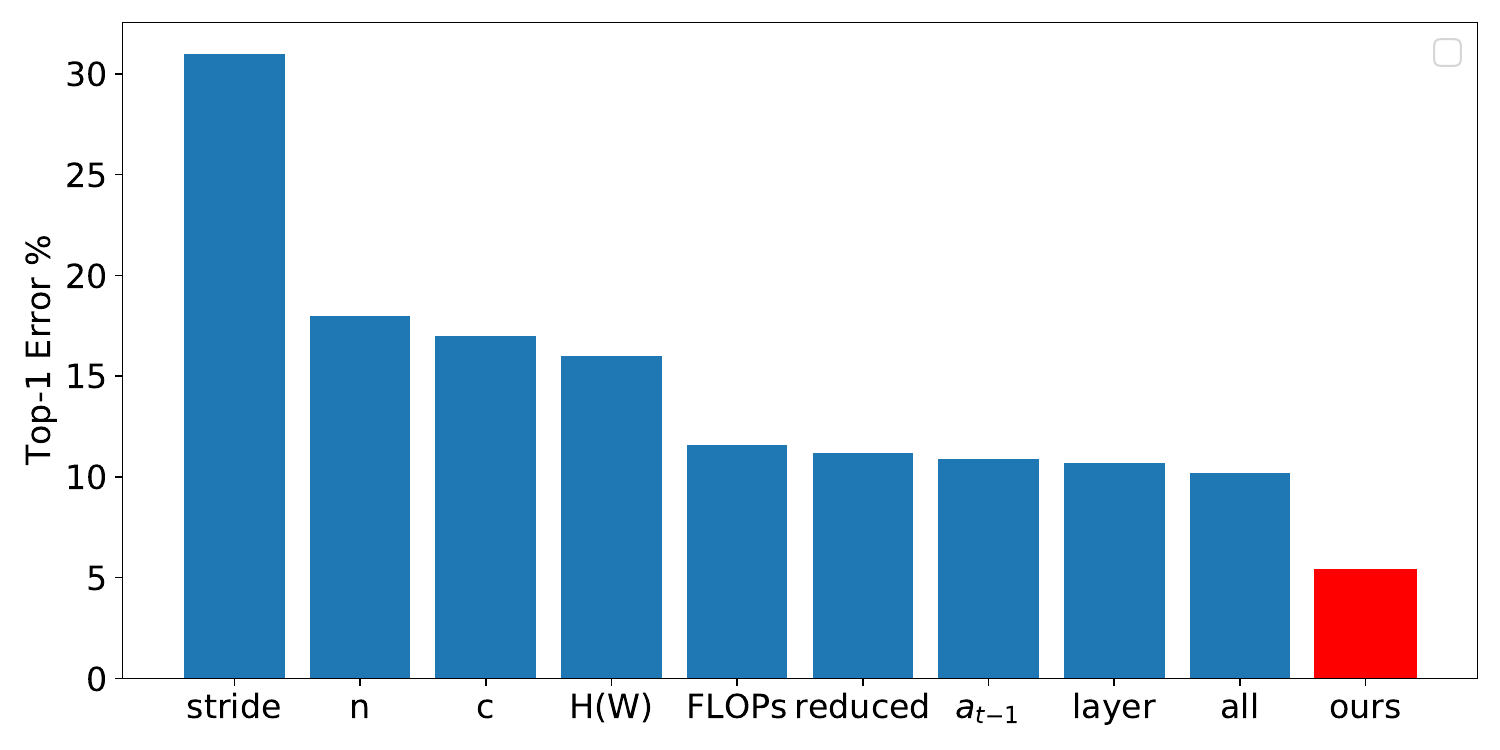}
\end{center}
   \caption{An error-rate comparison for individual AMC layer embedding, overall AMC, and AGMC layer embedding. Our method has achieved roughly 2$\times$ less error rate.}
\label{fig:layer_embedding}
\end{figure}

%% file: Figure_5.tex
\begin{figure}[t]
\begin{center}
   \includegraphics[width=1\linewidth]{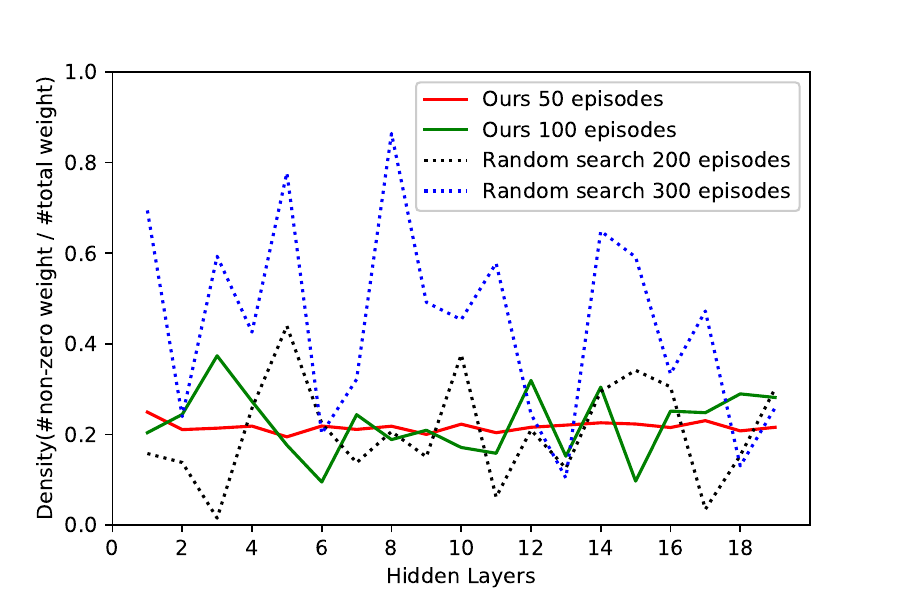}
\end{center}
   \caption{Comparing AGMC pruning stability across different layers with random search on ResNet-20. Random search uses 200 and 300 episodes, achieving a compressed network with $71\%$ and $88.41\%$ validation accuracy, respectively. AGMC searches for 50 episodes and 100 episodes with a validation accuracy of $93.8\%$ and $94.6\%$, respectively. Thus, AGMC achieved a higher compression ratio with considerably fewer episodes.}
\label{fig:random_search}
\end{figure}

%% file: Figure_Acc_FLOPs.tex
\begin{figure}[t]
\begin{center}
   \includegraphics[width=1\linewidth]{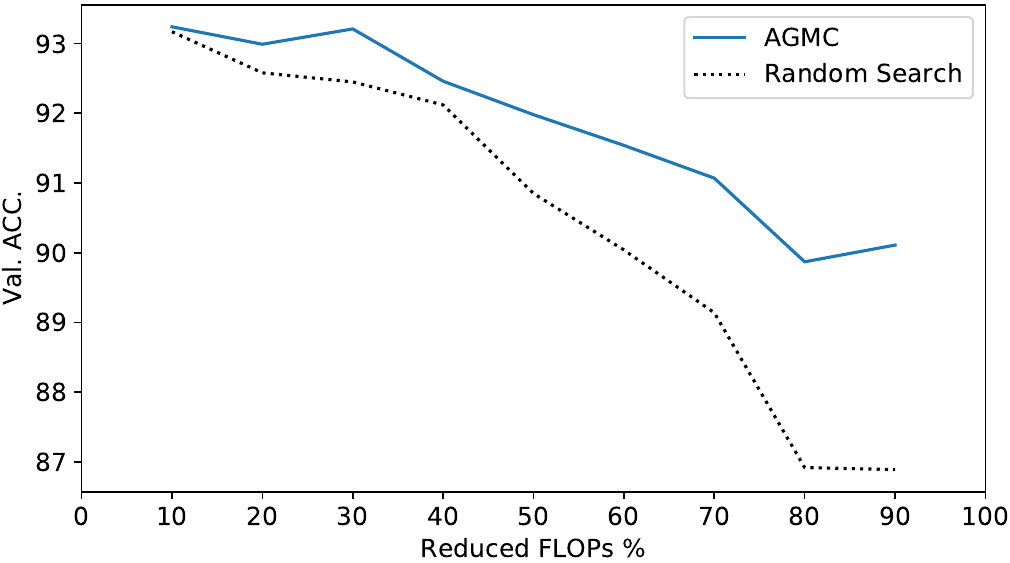}
\end{center}
   \caption{Validation accuracy comparison of random search and AGMC on ResNet-56 under different FLOPs constraints.}
\label{fig:Acc_FLOPs}
\end{figure}

%% file: Table_1.tex
\begin{table}
\caption{Pruning policy comparison of FLOPs-constrained compression on ResNet-$20/32/44/56/110$ and VGG-16~\cite{Simonyan2015VGG}. The ResNet family are trained on CIFAR-10 and VGG-16 is trained on the   ImageNet~(ILSVRC-2012) dataset.}
\begin{center}
\begin{tabular}{|l|c|c|c|c|}
\hline
Model & Method & FLOPs & \makecell{Test \\$Acc.\%$} & $\Delta Acc.$\\
\hline\hline
\multirow{5}{*}{ResNet20}   &Deep   & 50\%  &  79.6 & -12.13 \\
                            &Shallow& 50\%  &  83.2 & -8.53 \\
                            &Uniform& 50\%  &  84 & -7.73 \\
                            &SSL    & 52\%  &  89.78& -2.39\\
                            &MorphNet& 48\%&90.1 &-2.07\\
                            &Rethink& 60\%& 91.07& -1.34\\
                            &SFP& 58\%& 90.83& -1.37\\
                            &DSA& 50\%&91.38 & -0.79\\
                            &AMC    & 50\%  &  86.4 & -5.33 \\
                            &\textbf{AGMC}   & 50\%  &  \textbf{91.42} & \textbf{-0.31} \\

\hline\hline

\multirow{5}{*}{ResNet56}   &Uniform& 50\%  & 87.5 & -5.89 \\
                            &Deep   & 50\%  & 88.4 & -4.99 \\
                            &SSL    & 47\% &91.22 &-1.90\\
                            &MorphNet& 52\%&91.55 &-1.57\\
                            &Rethink & 50\%&93.07 &-0.73\\
                            &SFP& 50\%& 92.26& -1.33\\
                            &AMC    & 50\%  & 90.2 & -3.19 \\
                            &\textbf{AGMC}   & 50\%  & 92.76 & \textbf{-0.63}\\
\hline\hline

\multirow{4}{*}{VGG-16}     &FP & 20\%    & 55.9 & -14.6 \\
                            &RNP & 20\%    & 66.92 & -3.58 \\
                            &SPP & 20\%    & 68.2 & -2.3 \\
                            &AMC & 20\%    & 69.1 & -1.4 \\
                            &\textbf{AGMC} & 20\%    & \textbf{70.35} & \textbf{-0.15} \\
\hline\hline

\multirow{2}{*}{ResNet110}  &RS     & 50\%  &  87.26    &  -6.42 \\
                            &\textbf{AGMC}   & 50\%  & \textbf{93.08}     & \textbf{-0.6} \\
                            
\hline\hline

\multirow{2}{*}{ResNet44}   &RS     & 50\%  &   88.14 & -4.96  \\
                            &\textbf{AGMC}   & 50\%  &  \textbf{92.28} & \textbf{-0.82}  \\
                            
\hline\hline
\multirow{2}{*}{ResNet32}   &RS     & 50\%  &   89.57 & -3.06 \\
                            &\textbf{AGMC}   & 50\%  &  \textbf{90.96} & \textbf{-1.67} \\

\hline

\end{tabular}
\end{center}

\label{table:1}
\vspace{-10pt}
\end{table}

%% file: Table_2.tex
\begin{table}
\caption{Pruning policy comparison of FLOPs-constrained compression on MobileNet-v1/v2 and ShuffleNet-v1/v2. The MobileNet-v1/v2 are trained on the ImageNet~(ILSVRC-2012) dataset, and the ShuffleNet-v1/v2 are trained on the CIFAR-100 dataset. The column FLOPs denotes the ratio between the FLOPs of the compressed model and the original model.}
\begin{center}
\small

\begin{tabular}{|l|c|c|p{9mm}|p{9mm}|}
\hline
Model & Method &FLOPs & Test \newline Acc.&$\Delta Acc. \newline \%$  \\

\hline\hline
\multirow{4}{*}{MobileNet-v1}   & uniform\cite{Andrew2017MobileNetv1}       & $56\%$  & 68.10 &$-2.5$\\
                                & {uniform\cite{Andrew2017MobileNetv1}}     & $41\%$  & 66.90 &-3.7\\
                                & {AMC\cite{he2018amc}}                     & $40\%$  &  68.90& -1.7\\
                                & \textbf{AGMC}                             & $\textbf{40\%}$   &\textbf{69.40} & \textbf{-1.2}\\

\hline\hline
\multirow{3}{*}{MobileNet-v2}   & uniform \cite{Sandler2018mobileNetv2}     & \multirow{3}{*}{$70\%$}  & 69.80&-2\\
                                & AMC\cite{he2018amc}                       &                          &  70.80 &-1\\
                                & \textbf{AGMC}                         &                              &\textbf{70.87}     & \textbf{-0.93} \\
\hline\hline
\multirow{2}{*}{ShuffleNet-v1}  
                                &RS                         &  $60\%$           &     63.70   & $ -4.94$ \\
                                &\textbf{AGMC}                       &  $60\%$       &     \textbf{65.26}       & $ \textbf{-3.38}$\\

\hline\hline
\multirow{2}{*}{ShuffleNet-v2} 
                                &RS                         &  $60\%$       &  65.74& $ -3.11$\\
                                &\textbf{AGMC}               & $60\%$  &    \textbf{66.28}    & $\textbf{-2.57} $ \\

\hline
\end{tabular}
\end{center}
\label{table:2}
\vspace{-10pt}
\end{table}

%% file: Table_3.tex
\begin{table}
\caption{Latency and GPU memory usage of pruned models using AGMC. We analyzed ResNet-20/56 on CIFAR-10 and VGG-16 and MobileNet-v1 on ImageNet dataset.}
\begin{center}
\begin{tabular}{|l|c|c|c|}
\hline
Model & FLOPs & Latency &GPU Mem. \\
\hline\hline
\multirow{2}{*}{MobileNet-v1}          & $100\%$FLOPs            & 11.02ms&       17MB \\
                                    & $40\%$FLOPs         &10.52ms& 14MB \\

\hline\hline

\multirow{2}{*}{VGG-16}          & $100\%$FLOPs            & 20.52ms&       528MB \\
                                    & $20\%$FLOPs         &16.82ms& 387MB \\
                                    
\hline\hline

\multirow{2}{*}{ResNet-56}          & $100\%$FLOPs            & 0.52ms&       3.4MB \\
                                    & $50\%$FLOPs         &0.48ms& 1.8M \\
    
\hline\hline
\multirow{2}{*}{ResNet-20}   & $100\%$FLOPs            & 0.32ms&       1.1MB \\
                                    & $50\%$FLOPs         &0.30ms& 565KB \\

\hline

\end{tabular}
\end{center}

\label{table:3}
\end{table}

%% file: 05_Conclusion.tex
\section{Conclusion}
This paper proposed an Auto Graph encoder-decoder Model Compression (AGMC), which combines graph convolutional networks and reinforcement learning to explore network compression policies automatically. To the best of our knowledge, this is the first work to model DNNs as computational graphs to enhance model compression. Furthermore, we conducted comprehensive experiments on over-parameterized and mobile-friendly DNNs. In the experiment, we show the superiority of our learning-based DNN embedding. By learning DNNs' embedding from their structure information, AGMC outperforms all the rule-based DNN embedding methods by a large margin. On over-parameterized, such as ResNet-56, our method defeats all the baselines with only 0.63\% accuracy loss.  Additionally, AGMC successfully compressed mobile-friendly DNNs, which are already compact. For instance, in the MobileNet-V1, we achieve a higher compression ratio than baselines with only 1.2\% accuracy loss.

%% file: egpaper_for_review.bbl
\begin{thebibliography}{10}\itemsep=-1pt

\bibitem{abadi2016tensorflow}
Martin Abadi, Paul Barham, Jianmin Chen, Zhifeng Chen, Andy Davis, Jeffrey
  Dean, Matthieu Devin, Sanjay Ghemawat, Geoffrey Irving, Michael Isard,
  Manjunath Kudlur, Josh Levenberg, Rajat Monga, Sherry Moore, Derek~G. Murray,
  Benoit Steiner, Paul Tucker, Vijay Vasudevan, Pete Warden, Martin Wicke, Yuan
  Yu, and Xiaoqiang Zheng.
\newblock Tensorflow: A system for large-scale machine learning.
\newblock In {\em 12th USENIX Symposium on Operating Systems Design and
  Implementation (OSDI 16)}, pages 265--283, 2016.

\bibitem{blakeney2020is}
Cody Blakeney, Yan Yan, and Ziliang Zong.
\newblock Is pruning compression?: Investigating pruning via network layer
  similarity.
\newblock {\em WACV}, pages 903--911, 2020.

\bibitem{ADMM}
Stephen Boyd, Neal Parikh, Eric Chu, Borja Peleato, and Jonathan Eckstein.
\newblock Distributed optimization and statistical learning via the alternating
  direction method of multipliers.
\newblock {\em Found. Trends Mach. Learn.}, 3(1):1–122, Jan. 2011.

\bibitem{chen2019renas}
Yukang Chen, Gaofeng Meng, Qian Zhang, Shiming Xiang, Chang Huang, Lisen Mu,
  and Xinggang Wang.
\newblock Renas - reinforced evolutionary neural architecture search.
\newblock {\em CVPR}, pages 4787--4796, 2019.

\bibitem{Dudziak2021BPR_NAS}
Lukasz Dudziak, Thomas Chau, Mohamed~S. Abdelfattah, Royson Lee, Hyeji Kim, and
  Nicholas~D. Lane.
\newblock {BRP-NAS}: Prediction-based {NAS} using gcns.
\newblock 2021.

\bibitem{gao2019graphnas}
Yang Gao, Hong Yang, Peng Zhang, Chuan Zhou, and Yue Hu.
\newblock Graphnas: Graph neural architecture search with reinforcement
  learning.
\newblock {\em arXiv: Learning}, 2019.

\bibitem{gordon2018morphnet}
Ariel Gordon, Elad Eban, Ofir Nachum, Bo Chen, Hao Wu, Tien-Ju Yang, and Edward
  Choi.
\newblock {MorphNet}: {Fast} \& {Simple} {Resource}-{Constrained} {Structure}
  {Learning} of {Deep} {Networks}.
\newblock In {\em 2018 {IEEE}/{CVF} {Conference} on {Computer} {Vision} and
  {Pattern} {Recognition}}, pages 1586--1595, Salt Lake City, UT, June 2018.
  IEEE.

\bibitem{Guo2019NAS_NAT}
Yong Guo, Yin Zheng, Mingkui Tan, Qi Chen, Jian Chen, Peilin Zhao, and Junzhou
  Huang.
\newblock {NAT}: Neural architecture transformer for accurate and compact
  architectures.
\newblock In {\em Proc. of the Advances in Neural Information Processing
  Systems}, volume~32, pages 737--748. Curran Associates, Inc., 2019.

\bibitem{han2017ese}
Song Han, Junlong Kang, Huizi Mao, Yiming Hu, Xin Li, Yubin Li, Dongliang Xie,
  Hong Luo, Song Yao, Yu Wang, Huazhong Yang, and (Bill) J.~William Dally.
\newblock {ESE}: Efficient speech recognition engine with sparse {LSTM} on
  {FPGA}.
\newblock {\em Proc. of the ACM/SIGDA International Symposium on
  Field-Programmable Gate Arrays}, pages 75--84, 2017.

\bibitem{han2015deep}
Song Han, Huizi Mao, and J.~William Dally.
\newblock Deep compression: Compressing deep neural network with pruning,
  trained quantization and huffman coding.
\newblock {\em International conference on learning representations}, 2015.

\bibitem{hassibi1992second}
Babak Hassibi and G.~David Stork.
\newblock Second order derivatives for network pruning: Optimal brain surgeon.
\newblock {\em NIPS}, pages 164--171, 1992.

\bibitem{he2016deep}
Kaiming He, Xiangyu Zhang, Shaoqing Ren, and Jian Sun.
\newblock Deep residual learning for image recognition.
\newblock pages 770--778, 2016.

\bibitem{he2018sfp}
Yang He, Guoliang Kang, Xuanyi Dong, Yanwei Fu, and Yi Yang.
\newblock Soft {Filter} {Pruning} for {Accelerating} {Deep} {Convolutional}
  {Neural} {Networks}.
\newblock In {\em Proceedings of the {Twenty}-{Seventh} {International} {Joint}
  {Conference} on {Artificial} {Intelligence}}, pages 2234--2240, Stockholm,
  Sweden, July 2018. International Joint Conferences on Artificial Intelligence
  Organization.

\bibitem{he2018amc}
Yihui He, Ji Lin, Zhijian Liu, Hanrui Wang, Li-Jia Li, and Song Han.
\newblock Amc: Automl for model compression and acceleration on mobile devices.
\newblock In {\em Proc. of the European Conference on Computer Vision (ECCV)},
  pages 784--800, 2018.

\bibitem{he2017channel}
Yihui He, Xiangyu Zhang, and Jian Sun.
\newblock Channel pruning for accelerating very deep neural networks.
\newblock In {\em Proc. of the IEEE International Conference on Computer
  Vision}, pages 1389--1397, 2017.

\bibitem{hinton2015distilling}
E.~Geoffrey Hinton, Oriol Vinyals, and Jeffrey Dean.
\newblock Distilling the knowledge in a neural network.
\newblock {\em CoRR}, 2015.

\bibitem{Andrew2017MobileNetv1}
Andrew~G. Howard, Menglong Zhu, Bo Chen, Dmitry Kalenichenko, Weijun Wang,
  Tobias Weyand, Marco Andreetto, and Hartwig Adam.
\newblock Mobilenets: Efficient convolutional neural networks for mobile vision
  applications.
\newblock {\em CoRR}, abs/1704.04861, 2017.

\bibitem{huang2018condensenet}
Gao Huang, Shichen Liu, Laurens Van~der Maaten, and Kilian~Q Weinberger.
\newblock Condensenet: An efficient densenet using learned group convolutions.
\newblock In {\em Proc. of the IEEE conference on computer vision and pattern
  recognition}, pages 2752--2761, 2018.

\bibitem{jacob2018quantization}
Benoit Jacob, Skirmantas Kligys, Bo Chen, Menglong Zhu, Matthew Tang, Andrew
  Howard, Hartwig Adam, and Dmitry Kalenichenko.
\newblock Quantization and training of neural networks for efficient
  integer-arithmetic-only inference.
\newblock In {\em Proc. of the IEEE Conference on Computer Vision and Pattern
  Recognition}, pages 2704--2713, 2018.

\bibitem{ji2018recom}
Houxiang Ji, Linghao Song, Li Jiang, Hai~Halen Li, and Yiran Chen.
\newblock Recom: An efficient resistive accelerator for compressed deep neural
  networks.
\newblock In {\em Proc of. Design, Automation \& Test in Europe Conference \&
  Exhibition (DATE)}, pages 237--240. IEEE, 2018.

\bibitem{article}
Thomas~N Kipf and Max Welling.
\newblock Semi-supervised classification with graph convolutional networks.
\newblock {\em arXiv preprint arXiv:1609.02907}, 2016.

\bibitem{kipf2017gcn}
Thomas~N. Kipf and Max Welling.
\newblock Semi-supervised classification with graph convolutional networks.
\newblock In {\em Proc. of the International Conference on Learning
  Representations (ICLR)}, 2017.

\bibitem{krizhevsky2009cifar}
A Krizhevsky and G Hinton.
\newblock Learning multiple layers of features from tiny images.
\newblock 2009.

\bibitem{FP_li2017pruning}
Hao Li, Asim Kadav, Igor Durdanovic, Hanan Samet, and Hans~Peter Graf.
\newblock Pruning filters for efficient convnets.
\newblock {\em CoRR}, abs/1608.08710, 2016.

\bibitem{lillicrap2015continuous}
Timothy~P Lillicrap, Jonathan~J Hunt, Alexander Pritzel, Nicolas Heess, Tom
  Erez, Yuval Tassa, David Silver, and Daan Wierstra.
\newblock Continuous control with deep reinforcement learning.
\newblock {\em arXiv preprint arXiv:1509.02971}, 2015.

\bibitem{Lin2017RNP}
Ji Lin, Yongming Rao, Jiwen Lu, and Jie Zhou.
\newblock Runtime neural pruning.
\newblock In {\em Proc. of the Advances in Neural Information Processing
  Systems}, pages 2181--2191, 2017.

\bibitem{liu2020autocompress}
Ning Liu, Xiaolong Ma, Zhiyuan Xu, Yanzhi Wang, Jian Tang, and Jieping Ye.
\newblock Autocompress: An automatic dnn structured pruning framework for
  ultra-high compression rates.
\newblock In {\em Proc. of Artificial Intelligence Conference (AAAI)}, pages
  4876--4883, 2020.

\bibitem{liu2019rethinking}
Zhuang Liu, Mingjie Sun, Tinghui Zhou, Gao Huang, and Trevor Darrell.
\newblock Rethinking the value of network pruning.
\newblock {\em International Conference on Learning Representations (ICLR)},
  2019.

\bibitem{ma2018shufflenetv2}
Ningning Ma, Xiangyu Zhang, Hai-Tao Zheng, and Jian Sun.
\newblock Shufflenet v2: Practical guidelines for efficient cnn architecture
  design.
\newblock In {\em Proc. of the European conference on computer vision (ECCV)},
  pages 116--131, 2018.

\bibitem{mammadli2019effdnn}
Rahim Mammadli, Felix Wolf, and Ali Jannesari.
\newblock The art of getting deep neural networks in shape.
\newblock volume~15, pages 62:1--62:21, Jan. 2019.

\bibitem{mehta2020dicenet}
Sachin Mehta, Hannaneh Hajishirzi, and Mohammad Rastegari.
\newblock Dicenet: Dimension-wise convolutions for efficient networks.
\newblock {\em IEEE Transactions on Pattern Analysis and Machine Intelligence},
  2020.

\bibitem{molchanov2019importance}
Pavlo Molchanov, Arun Mallya, Stephen Tyree, Iuri Frosio, and Jan Kautz.
\newblock Importance estimation for neural network pruning.
\newblock In {\em Proceedings of the IEEE Conference on Computer Vision and
  Pattern Recognition}, pages 11264--11272, 2019.

\bibitem{vedaldi2020dsa}
Xuefei Ning, Tianchen Zhao, Wenshuo Li, Peng Lei, Yu Wang, and Huazhong Yang.
\newblock {DSA}: {More} {Efficient} {Budgeted} {Pruning} via {Differentiable}
  {Sparsity} {Allocation}.
\newblock In Andrea Vedaldi, Horst Bischof, Thomas Brox, and Jan-Michael Frahm,
  editors, {\em European {Conference} on {Computer} {Vision} – {ECCV} 2020},
  volume 12348, pages 592--607, Cham, 2020. Springer International Publishing.
\newblock Series Title: Lecture Notes in Computer Science.

\bibitem{park2019relational}
Wonpyo Park, Dongju Kim, Yan Lu, and Minsu Cho.
\newblock Relational knowledge distillation.
\newblock In {\em Proc. of the IEEE Conference on Computer Vision and Pattern
  Recognition}, pages 3967--3976, 2019.

\bibitem{paszke2019pytorch}
Adam Paszke, Sam Gross, Francisco Massa, Adam Lerer, James Bradbury, Gregory
  Chanan, Trevor Killeen, Zeming Lin, Natalia Gimelshein, Luca Antiga, Alban
  Desmaison, Andreas Kopf, Edward Yang, Zachary DeVito, Martin Raison, Alykhan
  Tejani, Sasank Chilamkurthy, Benoit Steiner, Lu Fang, Junjie Bai, and Soumith
  Chintala.
\newblock Pytorch: An imperative style, high-performance deep learning library.
\newblock In H. Wallach, H. Larochelle, A. Beygelzimer, F. d\textquotesingle
  Alch\'{e}-Buc, E. Fox, and R. Garnett, editors, {\em Advances in Neural
  Information Processing Systems}, volume~32, pages 8026--8037. Curran
  Associates, Inc., 2019.

\bibitem{polino2018model}
Antonio Polino, Razvan Pascanu, and Dan Alistarh.
\newblock Model compression via distillation and quantization.
\newblock {\em ICLR}, 2018.

\bibitem{olga2015ImageNet}
Olga Russakovsky, Jia Deng, Hao Su, Jonathan Krause, Sanjeev Satheesh, Sean Ma,
  Zhiheng Huang, Andrej Karpathy, Aditya Khosla, Michael Bernstein,
  Alexander~C. Berg, and Li Fei-Fei.
\newblock {ImageNet Large Scale Visual Recognition Challenge}.
\newblock {\em International Journal of Computer Vision (IJCV)},
  115(3):211--252, 2015.

\bibitem{2013low-rank}
T.~N. {Sainath}, B. {Kingsbury}, V. {Sindhwani}, E. {Arisoy}, and B.
  {Ramabhadran}.
\newblock Low-rank matrix factorization for deep neural network training with
  high-dimensional output targets.
\newblock In {\em 2013 IEEE International Conference on Acoustics, Speech and
  Signal Processing}, pages 6655--6659, 2013.

\bibitem{Sandler2018mobileNetv2}
Mark Sandler, G.~Andrew Howard, Menglong Zhu, Andrey Zhmoginov, and Liang-Chieh
  Chen.
\newblock {MobileNetV2}: Inverted residuals and linear bottlenecks.
\newblock pages 4510--4520, 2018.

\bibitem{Schlichtkrull2018rgcn}
Michael Schlichtkrull, Thomas~N. Kipf, Peter Bloem, Rianne van den Berg, Ivan
  Titov, and Max Welling.
\newblock Modeling relational data with graph convolutional networks.
\newblock In Aldo Gangemi, Roberto Navigli, Maria-Esther Vidal, Pascal Hitzler,
  Rapha{\"e}l Troncy, Laura Hollink, Anna Tordai, and Mehwish Alam, editors,
  {\em The Semantic Web}, pages 593--607, Cham, 2018. Springer International
  Publishing.

\bibitem{schulman2017ppo}
John Schulman, Filip Wolski, Prafulla Dhariwal, Alec Radford, and Oleg Klimov.
\newblock Proximal policy optimization algorithms, 2017.

\bibitem{Han2020NAS_oneshot}
Han Shi, Renjie Pi, Hang Xu, Zhenguo Li, James~T. Kwok, and Tong Zhang.
\newblock Bridging the gap between sample-based and one-shot neural
  architecture search with {BONAS}.
\newblock 2020.

\bibitem{DPG}
David Silver, Guy Lever, Nicolas Heess, Thomas Degris, Daan Wierstra, and
  Martin Riedmiller.
\newblock Deterministic policy gradient algorithms.
\newblock volume~1, 06 2014.

\bibitem{Simonyan2015VGG}
Karen Simonyan and Andrew Zisserman.
\newblock Very deep convolutional networks for large-scale image recognition.
\newblock {\em international conference on learning representations}, 2015.

\bibitem{sutton1999policy}
S.~Richard Sutton, A.~David Mcallester, P.~Satinder Singh, and Yishay Mansour.
\newblock Policy gradient methods for reinforcement learning with function
  approximation.
\newblock {\em Neural Information Processing Systems}, pages 1057--1063, 1999.

\bibitem{swaminathan2020sparse}
Sridhar Swaminathan, Deepak Garg, Rajkumar Kannan, and Frederic Andres.
\newblock Sparse low rank factorization for deep neural network compression.
\newblock {\em Neurocomputing}, pages 185--196, 2020.

\bibitem{LSTM}
Sheng~Kai Tai, Richard Socher, and D.~Christopher Manning.
\newblock Improved semantic representations from tree-structured long
  short-term memory networks.
\newblock {\em International Workshop on the ACL2 Theorem Prover and Its
  Applications}, 2015.

\bibitem{wang2017SPP}
Huan Wang, Qiming Zhang, Yuehai Wang, and Roland Hu.
\newblock Structured probabilistic pruning for deep convolutional neural
  network acceleration.
\newblock {\em British Machine Vision Conference}, 2017.

\bibitem{wang2020bamc}
J. {Wang}, H. {Bai}, J. {Wu}, and J. {Cheng}.
\newblock Bayesian automatic model compression.
\newblock {\em IEEE Journal of Selected Topics in Signal Processing},
  14(4):727--736, 2020.

\bibitem{wen2016ssl}
Wei Wen, Chunpeng Wu, Yandan Wang, Yiran Chen, and Hai Li.
\newblock Learning {Structured} {Sparsity} in {Deep} {Neural} {Networks}.
\newblock In {\em Advances in {Neural} {Information} {Processing} {Systems}},
  volume~29. Curran Associates, Inc., 2016.

\bibitem{xiao2019autoprun}
Xia Xiao, Zigeng Wang, and Sanguthevar Rajasekaran.
\newblock Autoprun: Automatic network pruning by regularizing auxiliary
  parameters.
\newblock {\em Advances in Neural Information Processing Systems (NIPS)}, pages
  13681--13691, 2019.

\bibitem{xu2019how}
Keyulu Xu, Weihua Hu, Jure Leskovec, and Stefanie Jegelka.
\newblock How powerful are graph neural networks?
\newblock In {\em Proc. of International Conference on Learning Representations
  (ICLR)}, 2019.

\bibitem{NEURIPS2018_53f0d7c5}
Muhan Zhang and Yixin Chen.
\newblock Link prediction based on graph neural networks.
\newblock In {\em Advances in Neural Information Processing Systems}, pages
  5165--5175, 2018.

\bibitem{zhang2018shufflenet}
Xiangyu Zhang, Xinyu Zhou, Mengxiao Lin, and Jian Sun.
\newblock Shufflenet: An extremely efficient convolutional neural network for
  mobile devices.
\newblock In {\em Proc. of the IEEE conference on computer vision and pattern
  recognition}, pages 6848--6856, 2018.

\bibitem{zoph2017neural}
Barret Zoph and V.~Quoc Le.
\newblock Neural architecture search with reinforcement learning.
\newblock {\em international conference on learning representations}, 2017.

\end{thebibliography}
